\documentclass[letterpaper,10 pt,conference]{ieeeconf}
\IEEEoverridecommandlockouts                              

\usepackage{amsmath}

\usepackage{pdfpages}

\usepackage{graphicx}

\begin{document}
%
\title{\LARGE \bf TeraSim: Uncovering Unknown Unsafe Events for Autonomous Vehicles through Generative Simulation}
%
%
%

\author{Haowei Sun\textsuperscript{1†}, Xintao Yan\textsuperscript{1†}, Zhijie Qiao\textsuperscript{1†}, Haojie Zhu\textsuperscript{1†}, Yihao Sun\textsuperscript{1}, 
Jiawei Wang\textsuperscript{1},\\
Shengyin Shen\textsuperscript{2}, Darian Hogue\textsuperscript{2}, Rajanikant Ananta\textsuperscript{2},  Derek Johnson\textsuperscript{2}, Greg Stevens\textsuperscript{2}, \\Greg McGuire\textsuperscript{2}, 
Yifan Wei\textsuperscript{3},
Wei Zheng\textsuperscript{3},
Yong Sun\textsuperscript{3},
Yasuo Fukai\textsuperscript{3},
Henry X. Liu\textsuperscript{1,2*}%
\thanks{This research was partially funded by the United States National Science Foundation through the Mcity 2.0 Project (CMMI $\#2223517$).}
\thanks{\textsuperscript{1}Department of Civil and Environmental Engineering, University of Michigan, Ann Arbor, MI, 48109.}%
\thanks{\textsuperscript{2}University of Michigan Transportation
Research Institute, Ann Arbor, MI, 48109.}%
\thanks{\textsuperscript{3}ISUZU Technical Center of America, Inc., 46401 Commerce Center Dr,
 Plymouth, MI 48170.}%
\thanks{†These authors contributed equally to this work.}%
\thanks{*Corresponding author: Henry X. Liu (henryliu@umich.edu).}%
}


\maketitle

\begin{abstract}

Traffic simulation is essential for autonomous vehicle (AV) development, enabling comprehensive safety evaluation across diverse driving conditions. However, traditional rule-based simulators struggle to capture complex human interactions, while data-driven approaches often fail to maintain long-term behavioral realism or generate diverse safety-critical events. To address these challenges, we propose \textit{TeraSim}, an open-source, high-fidelity traffic simulation platform designed to uncover unknown unsafe events and efficiently estimate AV statistical performance metrics, such as crash rates. TeraSim is designed for seamless integration with third-party physics simulators and standalone AV stacks, to construct a complete AV simulation system. Experimental results demonstrate its effectiveness in generating diverse safety-critical events involving both static and dynamic agents, identifying hidden deficiencies in AV systems, and enabling statistical performance evaluation. These findings highlight TeraSim’s potential as a practical tool for AV safety assessment, benefiting researchers, developers, and policymakers. The code is available at https://github.com/mcity/TeraSim.

\end{abstract}


%
\IEEEpeerreviewmaketitle





\vspace{0.1cm}

\section{Introduction}

A comprehensive simulation system is essential for autonomous vehicle (AV) development. Simulation provides a virtual environment where AVs can encounter diverse and potentially hazardous scenarios in a controllable, scalable, and cost-effective manner. Unlike real-world testing, simulation allows for rapid iteration and testing without physical risks or high costs. Furthermore, simulation systems can generate synthetic data to complement real-world datasets, enriching training and validation processes to improve AV safety.

A robust AV simulation system consists of three key components: a \textit{traffic simulator}, a \textit{physics simulator}, and an \textit{AV stack}. The traffic simulator is responsible for generating the microscopic behavior of traffic participants, such as vehicles, pedestrians, and cyclists, creating an interactive environment for AV testing. Traditional model-based traffic simulators include systems like SUMO \cite{lopez2018microscopic-SUMO}, while recent advancements include data-driven approaches like NeuralNDE \cite{yan2023learning-NNDE} and SMART \cite{wu2025-smart}. The physics simulator handles sensor simulation, vehicle dynamics, and digital assets within the simulation environment. Sensor simulation provides data inputs to the AV stack, such as photorealistic renderings and LiDAR point clouds. Notable physics simulators, such as CARLA \cite{dosovitskiy2017-CARLA}, are widely used in research and development. Although some physics simulators include traffic simulation functionalities, the behavior of traffic agents remains unrealistic, lacking the naturalistic and statistical characteristics of real-world traffic. The AV stack represents the AV software system that interacts with the simulation environment, performing perception, decision-making, and control tasks. Open-source examples include Autoware \cite{autoware}, a ROS2-based autonomous driving framework, and Apollo \cite{baidu_apollo}, a full-stack platform supporting various autonomy levels. Seamless integration of these three components is critical to enable comprehensive AV development, testing, and validation.

Among these components, the traffic simulator plays a crucial role, as the behavioral realism of simulated agents directly impacts the reliability and effectiveness of simulation results. Therefore, accurate modeling of human agent behavior is essential to replicate real-world traffic environments with high fidelity. Additionally, to enhance AV performance effectively, the simulation system should generate valuable data, particularly safety-critical events. By systematically exposing AVs to diverse long-tail safety-critical conditions, the simulation system helps identify and address unknown unsafe conditions, ultimately improving AV robustness and reliability.

To address these needs, we introduce \textit{TeraSim}, an open-source, high-fidelity traffic simulation platform designed specifically for AV development. TeraSim employs a generative paradigm to construct traffic environments using two core models. The Naturalistic Driving Environment (NDE) reconstructs realistic traffic conditions with statistical fidelity, ensuring that generated environments follow real-world distributions. Building on NDE, the Naturalistic and Adversarial Driving Environment (NADE) amplifies rare but critical events, systematically exposing AVs to unknown unsafe conditions that existing traffic simulators often fail to capture. 

TeraSim offers three key contributions. First, it helps uncover unknown unsafe events through generative simulation, systematically exposing AVs to rare but critical driving conditions. This capability enhances the effectiveness of AV testing by uncovering hidden failure modes. 
Second, TeraSim provides a quantitative framework for estimating AV crash rates, enabling statistically grounded safety assessments that extend beyond predefined scenarios and support rigorous, data-driven validation. Third, TeraSim is designed for seamless integration, featuring an API-driven architecture that facilitates real-time integration with physics simulators and AV stacks, enabling co-simulation (Co-Sim) and leveraging advancements across the AV research community. By offering a scalable, high-fidelity simulation platform, TeraSim aims to accelerate the development of safer AV systems.

The rest of this paper is structured as follows. Section II reviews related work. Section III describes the architecture and functionalities of TeraSim. Section IV outlines the key advantages of TeraSim over existing traffic simulators. Section V demonstrates its application in building a comprehensive AV testing simulation system. Finally, Section VI concludes the paper.

\vspace{0.1cm}

\section{Related Work}

\subsection{Traffic Simulation}

Traditional traffic simulators such as SUMO \cite{lopez2018microscopic-SUMO} rely on rule-based models like IDM \cite{treiber2000congested-IDM} and MOBIL \cite{kesting2007general-MOBIL} to characterize human driving behavior. However, these models have limited representational power and cannot fully capture the complexity of human interactions. Recent advancements leverage neural networks to model multi-agent interactions, improving behavioral realism. For example, TrafficSim \cite{suo2021-trafficsim} introduced a data-driven approach that learns from real-world behavior, NeuralNDE \cite{yan2023learning-NNDE} proposed a generative framework that simulates traffic environments with statistical realism, capable of generating realistic safety-critical events, including crashes and near-misses. LimSim \cite{wen2023-limsim} enabled long-term interactive urban traffic simulations, while TrafficBots \cite{zhang2023-trafficbots} introduced scalable multi-agent world models.

More recently, LimSim++ \cite{fu2024-limsim++} integrated multi-modal large language models for enhanced closed-loop simulations, and ProSim \cite{tan2024-prosim} introduced a promptable system that allows fine-grained agent control using numerical, categorical, and textual inputs. LCSim \cite{zhang2024-lcsim} advanced large-scale traffic simulation by combining high-precision maps with diffusion models to generate realistic micro-traffic flow, and SMART \cite{wu2025-smart} proposed a tokenization scheme for modeling vectorized road networks and agent trajectories, achieving state-of-the-art performance on the WOMD Sim Agent Challenge \cite{montali2023-WOSAC}. Despite these advancements, most methods are limited to short time horizons (e.g., 8 seconds) or specific regions and lack the capability to comprehensively and systematically generate safety-critical events.

\subsection{Physics Simulation}

Physics simulators focus on sensor simulation and vehicle dynamics. CARLA \cite{dosovitskiy2017-CARLA}, an open-source platform built on the Unreal Engine, is widely used in both academia and industry, providing comprehensive sensor simulations and vehicle dynamics models. MetaDrive \cite{li2022-metadrive} provides a open-source lightweight environment for diverse driving scenarios. Other commercial solutions, including NVIDIA Omniverse \cite{Nvidia-Omniverse}, Applied Intuition \cite{Applied-Intuition}, Cognata \cite{cognata}, dSPACE \cite{dSPACE}, and IPG CarMaker \cite{IPG-CarMaker}, offer advanced capabilities but are often closed-source. More information of physics simulators can be found in the recent survey \cite{silva2024realistic-survey}.

\subsection{Generative AI solutions}


Recent advancements in generative AI have introduced World Models that unify traffic and physics simulations by producing video visual outputs that simultaneously depict agent behavior and sensor information, such as camera images. GAIA-1 \cite{hu2023-gaia1}, developed by Wayve, is a generative AI model capable of producing realistic driving videos using video, text, and action inputs. It learns key driving scene concepts such as vehicles, pedestrians, and road layouts, offering fine-grained control over ego-vehicle behavior and scene features. MagicDrive \cite{gao2023-magicdrive} and DriveDreamer \cite{wang2024-drivedreamer} enhance driving scene generation by integrating high-definition maps and 3D bounding boxes, improving video quality and enabling precise control over driving environments. UniSim \cite{yang2023-unisim} is a general-purpose generative simulation framework designed to create high-fidelity driving scenarios, supporting the testing and validation of autonomous driving algorithms. Although these methods achieve visual realism, they often struggle to ensure behaviorally accurate agent interactions and physically plausible vehicle dynamics. As simulation durations increase, unrealistic behaviors and physical violations become more prevalent, limiting their applicability to short-term, scenario-based simulations.

\section{TeraSim}
\subsection{System Architecture}
\begin{figure*}[ht]
    \vspace{0.15cm}
    \centering
    \includegraphics[width=\textwidth]{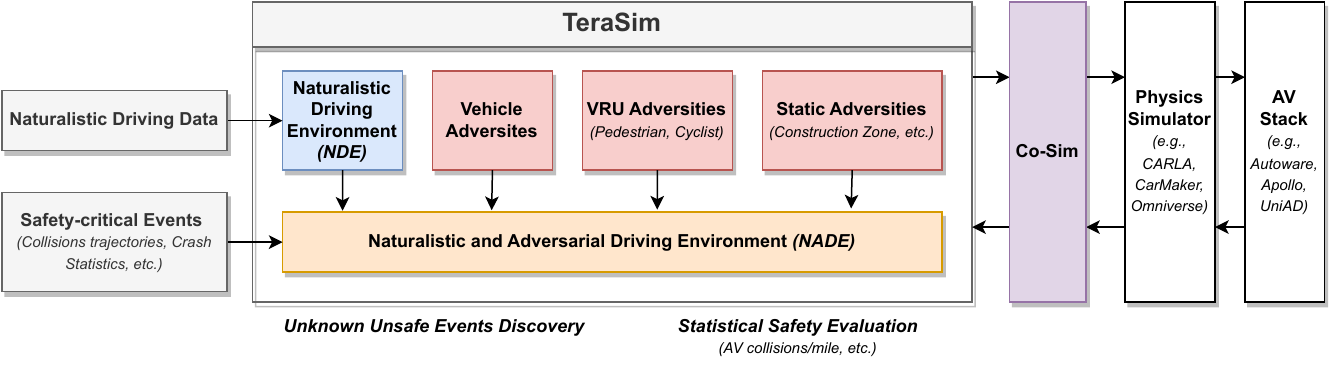}
    \caption{TeraSim Architecture.}
    \label{fig:architecture}
\end{figure*}

A detailed architectural diagram of TeraSim is shown in Fig. \ref{fig:architecture}. TeraSim is centered around NDE, which reconstructs realistic traffic conditions from naturalistic driving data. Built on top of NDE, adversities represent challenging or adversarial events. Each adversity is characterized by trigger conditions (defining when it occurs), trigger outcomes (describing its impact on the traffic environment), and trigger probabilities (determining its likelihood of occurrence).

The NADE further refines the simulation by leveraging safety-critical event data (e.g., collision trajectories, crash statistics) to intelligently regulate the frequency of adversities, ensuring a balanced exposure to rare but critical scenarios.  

TeraSim operates as a traffic simulator and connects to physics simulators through a Co-Sim module. At each simulation step, TeraSim provides traffic participant states to the physics simulator while receiving the AV state from it. Finally, the simulation pipeline integrates with the AV stack, enabling comprehensive testing and evaluation of autonomous driving systems. This framework enables the AV to navigate through dynamic traffic, uncovering unknown unsafe events missed in scenario-based simulations. Iterative testing accelerates AV safety evaluation, providing efficient crash rate quantification and failure mode identification.

\subsection{Traffic Environment}
\subsubsection{Environment Components}

In TeraSim, the traffic environment is composed of both static and dynamic components, each of which can be explicitly controlled to create diverse and realistic driving environments. The static components include environmental factors such as weather conditions, which influence road surface properties and visibility, as well as infrastructure elements like traffic lights and traffic signs that regulate vehicle and pedestrian movement. Additionally, variations in the static environment, such as construction zones or lane closures, introduce dynamic constraints that impact traffic flow and decision-making for AVs. These elements are predefined but can be systematically modified to study their effects on driving behaviors and AV performance.

The dynamic components consist of diverse traffic participants in the environment. The movement, reaction times, and interactive patterns of these agents can be controlled to introduce varying levels of complexity, ranging from routine urban traffic to highly adversarial environments. By adjusting these controllable parameters, TeraSim enables a structured exploration of traffic conditions, allowing for precise evaluation of AV safety and robustness in diverse driving environments.

\subsubsection{Naturalistic Driving Environment (NDE)}

NDE learns from large-scale naturalistic driving data to calibrate a realistic traffic behavior model that accurately represents real-world traffic dynamics. By integrating diverse traffic data, NDE reconstructs realistic traffic conditions with statistical fidelity, ensuring that agent behaviors align with observed human driving patterns. Within this environment, all traffic participants, including vehicles, pedestrians, and cyclists, operate interactively, responding dynamically to each other and to the tested AV. This interactive framework enables realistic traffic flow simulations, allowing AVs to be evaluated in a naturalistic yet controllable environment. A comprehensive description of the NDE modeling methodology can be found in our previous studies \cite{yan2023learning-NNDE, yan2021distributionally-NDE}.

\subsubsection{Adversity Orchestrator}

TeraSim enables the capability to insert adversity into any component of the environment, including both static and dynamic elements.  

\textbf{For static adversities}, TeraSim provides precise control over environmental conditions such as weather and traffic signals, allowing systematic variation in testing environments. Another key focus is the control over construction zones, which introduce significant disruptions to traffic flow and AV decision-making. TeraSim can dynamically generate construction zones of varying severity, ranging from partial lane closures to full road blockages, ensuring diverse and realistic constraints in the simulation.  

Since our NDE is fully interactive, the overall traffic environment naturally adapts to changes in static components. When adversities such as lane closures or road blockages are introduced, surrounding traffic participants dynamically adjust their behaviors in response, leading to emergent congestion, rerouting patterns, and altered interactive dynamics. This results in a more challenging and realistic testing environment for AVs, ensuring that they encounter diverse decision-making scenarios beyond predefined trajectories.

\textbf{For dynamic adversities}, traffic participants (vehicles, pedestrians, and cyclists) can randomly exhibit adversarial behaviors within the traffic environment, potentially leading to accidents or safety-critical events. To systematically define and manage these occurrences, we introduce a dedicated \textit{Dynamic Adversity Module}, which enables the controlled generation of such events. Each adversity is composed of three key components:  

\begin{itemize}

\item \textbf{Trigger Condition:}  
An adversity is activated based on a traffic participant’s perception of its environment, including its own position, surrounding vehicles, traffic lights, and other relevant contextual factors. For example, in a rear-end collision scenario, a possible trigger condition could be: if a vehicle detects another vehicle within \textit{20 meters ahead}, moving at a speed at least \textit{5 m/s slower} than itself, the rear-end adversity is activated. These conditions allow for a structured and flexible approach to defining when specific adversities should emerge.  

\item \textbf{Activated Behavior:}  
Once an adversity is triggered, it dictates how the affected traffic participant behaves. This can involve \textit{precisely controlled trajectories} or \textit{high-level behavioral commands} that lead to the emergence of hazardous situations. For instance, a pedestrian might suddenly jaywalk, a cyclist might swerve unpredictably, or a vehicle might fail to yield at an intersection. The activation mechanism ensures that adversarial behaviors are realistically represented and aligned with observed real-world driving risks. 

\item \textbf{Activation Probability:}  
Each adversity has an occurrence probability that reflects its \textit{real-world frequency}, balancing diversity while preventing an over-representation of low-probability events. This ensures that testing focuses on \textit{high-impact corner cases} rather than being diluted by statistically insignificant scenarios. Additionally, this probability can be \textit{calibrated dynamically} to align the overall accident rate and distribution with real-world crash data, maintaining both realism and efficiency in AV safety evaluations.  

\end{itemize}

By integrating these dynamically emerging adversities, TeraSim provides a \textit{highly adaptive and challenging testing environment} that closely mirrors real-world driving conditions, ensuring comprehensive AV validation.



\subsubsection{Naturalistic and Adversarial Driving Environment (NADE)}

NADE utilizes real-world safety-critical event data to calibrate the occurrence of different adversities within NDE, ensuring that the overall simulation environment maintains an accident rate consistent with real-world statistics. This calibration enhances the reliability of AV testing by aligning simulated risk exposure with real-world conditions.  

Additionally, NADE leverages dense deep reinforcement learning (D2RL) \cite{feng2023dense} to dynamically regulate the probability of different adversities emerging in the simulation. Instead of relying on fixed distributions, NADE continuously adjusts adversity frequency based on the evolving traffic state and AV interactions. By learning from safety-critical event data, NADE optimizes adversity occurrences to ensure a statistically representative distribution while exposing AVs to more diverse and high-risk scenarios. This approach not only uncovers rare failure modes but also accelerates real-world collision rate estimation by $10^3$ to $10^5$ times, enabling scalable and rigorous AV safety assessment. A detailed explanation can be found in \cite{feng2023dense, feng2021intelligent}.

\subsection{Co-Sim Design}

\begin{figure*}[ht]
    \centering
    \vspace{0.15cm}
    \includegraphics[width=\textwidth]{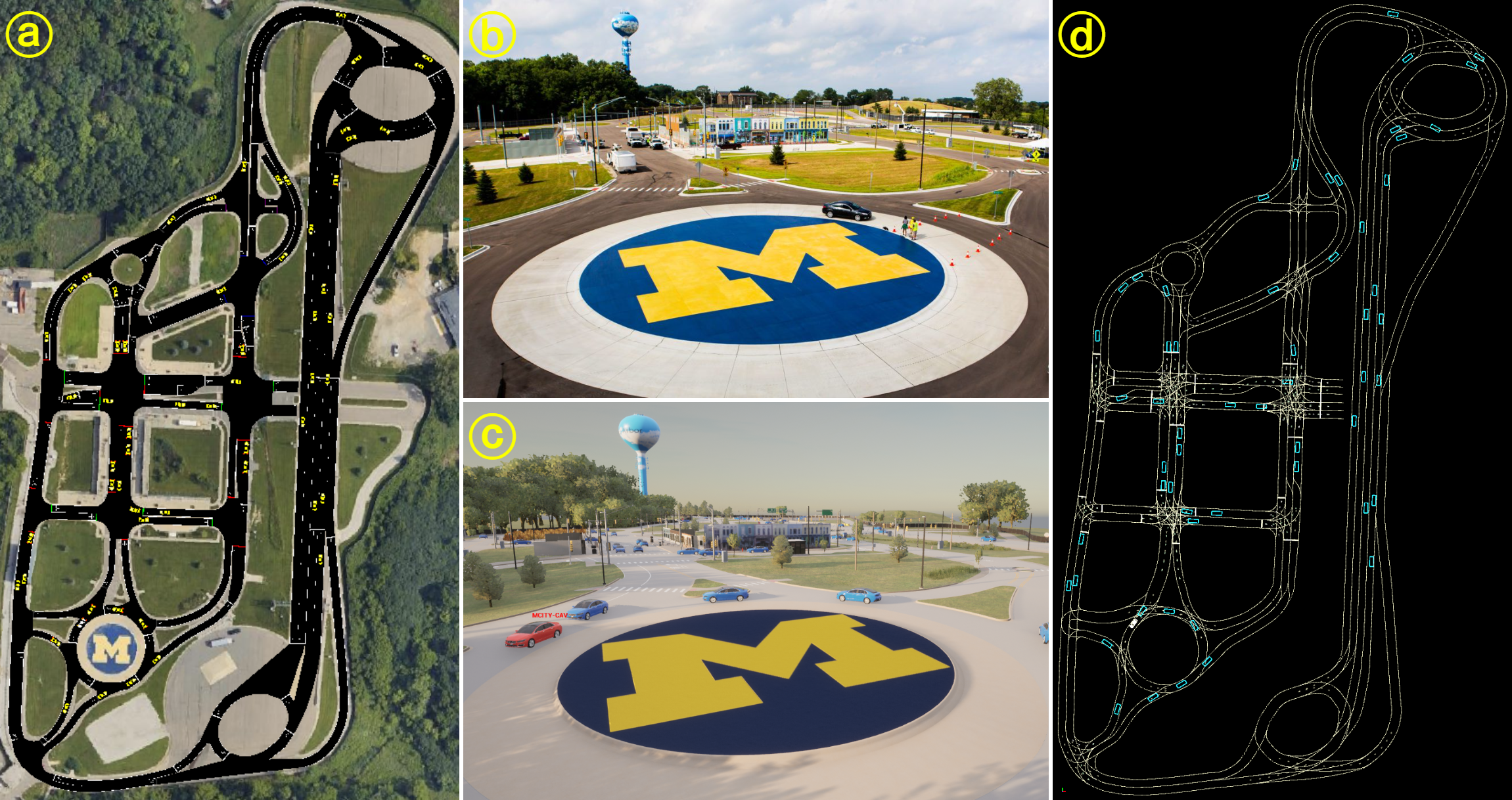}
    \caption{Panoramic view across different simulators with synchronized traffic. (a). TeraSim traffic generation. (b). Mcity real world. (c). Mcity CARLA digital twin. (d). Autoware Universe.}
    \label{fig:panoramic}
\end{figure*}

TeraSim seamlessly integrates with various physics simulators through a Co-Sim module, enabling it to exchange real-time traffic participant states while retrieving ego-vehicle dynamics from the physics engine. This flexible architecture allows for compatibility with different AV stacks, ensuring a comprehensive and adaptable testing framework for evaluating autonomous driving performance under diverse conditions.

The Co-Sim's real-time synchronization of traffic information across various platforms is facilitated through Redis \cite{Redis}. Redis is a high-performance in-memory data store that supports various communication patterns, including publish/subscribe, request/response, and GET/SET key. Operating over standard TCP/IP protocols, Redis typically exchanges data in JSON format, ensuring seamless integration across different systems and programming languages. Additionally, web-based tools like Redis Commander provide an intuitive interface for monitoring and managing Redis databases, enabling users to inspect stored keys and debug communication efficiently. A widely used architecture with a similar purpose is the Robot Operating System (ROS). Compared to ROS, our Redis-based implementation offers several advantages:

\begin{enumerate}
    \item Lightweight and Compatibility: Redis is lightweight, has minimal system dependencies, and supports almost all programming languages and operating systems, ensuring easy integration into existing workflows. In contrast, ROS is primarily designed for Ubuntu, supports only Python and C++, and its version-specific nature can introduce compatibility challenges.

    \item Cloud Support: Redis enables seamless remote access and can be deployed on cloud platforms like Amazon Web Services. With IP-based access, port configurations, and password authentication, multiple users can interact with the Co-Sim environment from different locations while running computationally intensive algorithms locally. In contrast, ROS has limited native cloud support and is typically confined to local networks.
\end{enumerate}

Co-Sim operates in a distributed, asynchronous mode, where each simulation platform runs its algorithms and manages data transmission at its own frequency. Data from each platform is sent to the Redis server as a JSON string under a unique key name, ensuring that each message contains platform-specific information managed exclusively by that platform. Messages may include various types of data, such as static traffic actors like traffic signs, traffic lights, and construction zones; dynamic traffic actors like vehicles, pedestrians, and cyclists; sensor data from LiDAR, Radar, Camera, IMU, and GNSS; and vehicle control commands such as throttle, brake, and steering. Each platform concurrently retrieves relevant data from other platforms to update its internal state. To prevent unintended cyclic updates, platforms must only modify their own traffic information.

To ensure consistency, all messages sent to Redis adhere to a predefined format, enabling robust data validation and error reporting. The data is structured hierarchically to reduce redundancy and clarify relationships. For instance, a general actor state message includes a header and a list of traffic actors, each specifying its ID, type, position, orientation, velocity, acceleration, and other key parameters. Every message undergoes automatic validation before transmission and reception, preventing a single corrupted message from compromising the entire system.

\subsection{TeraSim Workflow}

To illustrate, a fully integrated Co-Sim system combining a physics simulator and an AV stack in a closed-loop testing scenario could function as follows. The outlined steps serve as an example rather than a strict requirement.

\begin{enumerate}
    \item \textbf{AV Initialization:} The physics simulator initializes the AV at a predefined spawn position and shares its initial state using the key \texttt{av-state-info}.
    
    \item \textbf{Traffic Generation:} TeraSim monitors the AV’s state, generates a set of static and dynamic traffic participants, and transmits their data using the key \texttt{terasim-actor-info}.
    
    \item \textbf{Sensing Update:} The physics simulator incorporates this traffic data and generates sensor outputs relevant to the AV, sending them under the key \texttt{physics-sim-sensor-info}.
    
    \item \textbf{AV Stack Processing:} The AV stack processes the sensor data, executing perception and planning modules. It then transmits the computed control command via the key \texttt{av-control-info}.
    
    \item \textbf{Physics Execution:} The physics simulator applies the received control command, updates the AV’s state, and synchronizes with the Co-Sim environment.
    
    \item \textbf{Iterative Loop:} The process repeats continuously as all systems update asynchronously.
\end{enumerate}

This closed-loop process enables realistic AV development by continuously synchronizing simulated traffic, physics simulator, and AV behavior.

\vspace{0.2cm}
\section{TeraSim Advantages}

\vspace{0.1cm}

In this section, we discuss the advantages of TeraSim over existing traffic simulators in detail. 

\subsection{Uncovering Unknown Unsafe Events}

A key advantage of TeraSim is its ability to uncover unknown unsafe events through an interactive and adaptive testing environment. Unlike conventional traffic simulators that primarily model normal driving conditions or scenario-based testing methods that rely on fixed scenario sets, TeraSim dynamically generates diverse safety-critical scenarios, continuously challenging the AV under test. Its interactive nature ensures that surrounding agents respond in real-time to the AV’s behavior, systematically probing its safety performance.

With the NADE technique, TeraSim intensifies the generation of safety-critical events, significantly increasing the occurrence probability of long-tail, high-risk events. In particular, it enables the generation of sequential low-probability events within a time-continuous simulation environment (as demonstrated in Sec.~\ref{subsection:case-study}), making it more effective at revealing underlying deficiencies in AV system. In contrast, conventional methods require an extensive amount of time to encounter such rare events naturally, especially when evaluating a sequence of them.

Moreover, the unknown unsafe events identified by TeraSim inherently emerge during the testing process and are specific to each AV under test. Due to its interactive nature, TeraSim adaptively discovers unique failure modes for different AV systems, making it an intelligent and highly efficient evaluation framework. By shifting from a static, predefined scenario set to an adaptive and responsive simulation environment, TeraSim provides a more comprehensive and targeted safety assessment, uncovering hidden vulnerabilities that traditional approaches may struggle to reveal.

\subsection{Quantitative Safety Performance Evaluation}

Traditional scenario-based simulators struggle to estimate real-world crash rates because their traffic scenarios are predefined and do not reflect the statistical distributions observed in natural driving. The crash rate, defined as the number of crashes per total miles driven, is expressed as:

\[
R = \frac{N_{\text{crash}}}{D_{\text{total}}},
\]
where \( R \) represents the estimated crash rate, \( N_{\text{crash}} \) is the total number of crashes observed, and \( D_{\text{total}} \) is the total distance traveled by the AV during testing.

The accuracy of crash rate estimation relies on traffic environments being representative of real-world driving conditions. However, in scenario-based testing, the predefined cases fail to capture realistic traffic distributions, making crash rate estimates unreliable.

In contrast, our NDE framework reconstructs real-world traffic distributions, ensuring that AVs are tested in environments that statistically align with natural driving conditions. This enables meaningful and unbiased crash rate estimation, which scenario-based methods cannot achieve.

More importantly, NADE accelerates the generation of adversarial events by adjusting the probability of high-risk interactions for the AV under test. Therefore, TeraSim increases the frequency of crashes per mile driven, effectively reducing the total distance required for accurate crash rate estimation while maintaining statistical unbiasedness. This allows for scalable and rigorous AV safety evaluations in significantly less time than NDE testing.

\vspace{0.1cm}

\section{Demonstration}

\subsection{Overview}

In this section, we provide demonstrations and experimental results of TeraSim. First, we introduce the simulation platforms utilized in our testing and provide a panoramic view of the Co-Sim environment. Next, we analyze specific adversity events to highlight their nuanced complexities. Finally, we present a test case demonstrating how our design identifies potential issues in the AV system.

\subsection{Simulation Systems}

To evaluate our platform's performance, we developed a comprehensive simulation system for the Mcity testing facility—the world's first purpose-built facility for testing connected and automated vehicles. Our setup integrates TeraSim as the traffic simulator, CARLA as the physics simulator, and the Autoware Universe AV stack. Detailed information on these components is provided below.

\textbf{TeraSim} is our proposed traffic simulator that governs the behavior of both static and dynamic agents in the environment. In this study, we built it upon SUMO to leverage its low-level functionalities, including traffic signal control, road network modifications, and traffic flow generation. As a result, TeraSim offers robust simulation capabilities, generating diverse, long-tail safety-critical events within realistic traffic environments.

\textbf{CARLA} is an open-source simulator built with Unreal Engine, widely used by researchers for the development, training, and validation of AV systems. It offers realistic vehicle dynamics and a configurable sensor suite, including camera, LiDAR, radar, etc. The Mcity CARLA Digital Twin, developed and open-sourced by the Mcity team \cite{mcity_digital_twin}, is a virtual replica of the real-world Mcity test facility, created using digital assets within the CARLA simulator.

\textbf{Autoware} is an open-source software stack for autonomous driving research, development, and deployment. It integrates key functions like perception, localization, planning, and control. The latest version, Autoware Universe, enhances modularity, real-time performance, and production-level support. 

Fig. \ref{fig:panoramic} offers a comprehensive overview of the Co-Sim system in operation. Fig. \ref{fig:panoramic}a depicts the TeraSim environment with its generated traffic flows. Fig. \ref{fig:panoramic}b shows the Mcity real world. Fig. \ref{fig:panoramic}c illustrates the synchronized traffic flow in the Mcity CARLA digital twin, where the red Lincoln MKZ represents the AV. Finally, Fig. \ref{fig:panoramic}d presents the Autoware Universe bird’s-eye view, displaying a Lanelet2-format map alongside the full set of traffic agents. Collectively, these experiments underscore the flexibility and robustness of the Co-Sim system.

\subsection{Adversity Generation}

\begin{figure}[t]
    \vspace{0.15cm}
    \includegraphics[width=\columnwidth]{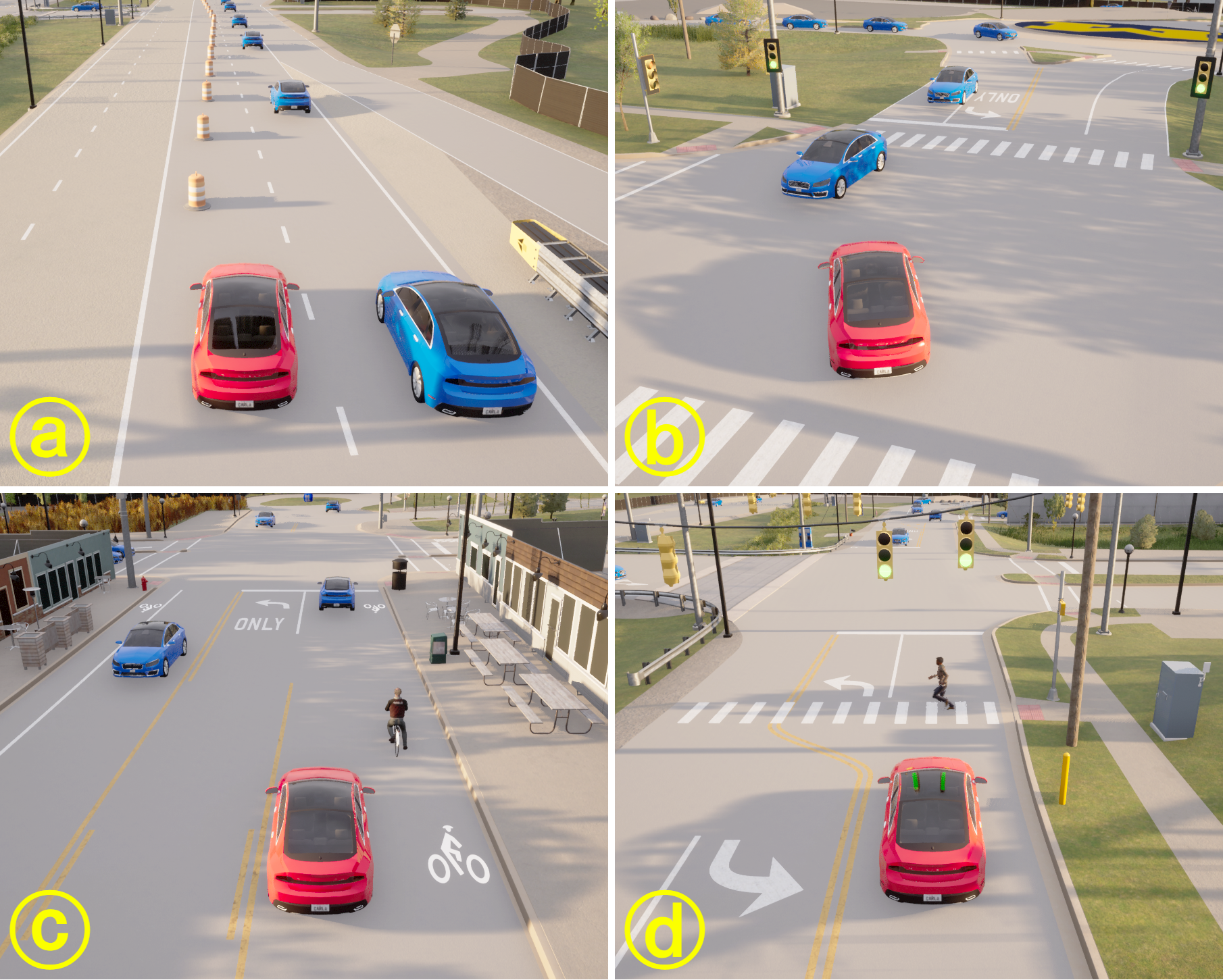}
    \caption{Example adversity generation. (a) Construction zone. (b) Vehicle adversity. (c) Cyclist adversity. (d) Pedestrian adversity.}
    \label{fig:adversity}
    \vspace{-0.15cm}
\end{figure}

\begin{figure*}[ht]
    \vspace{0.15cm}
    \includegraphics[width=\textwidth]{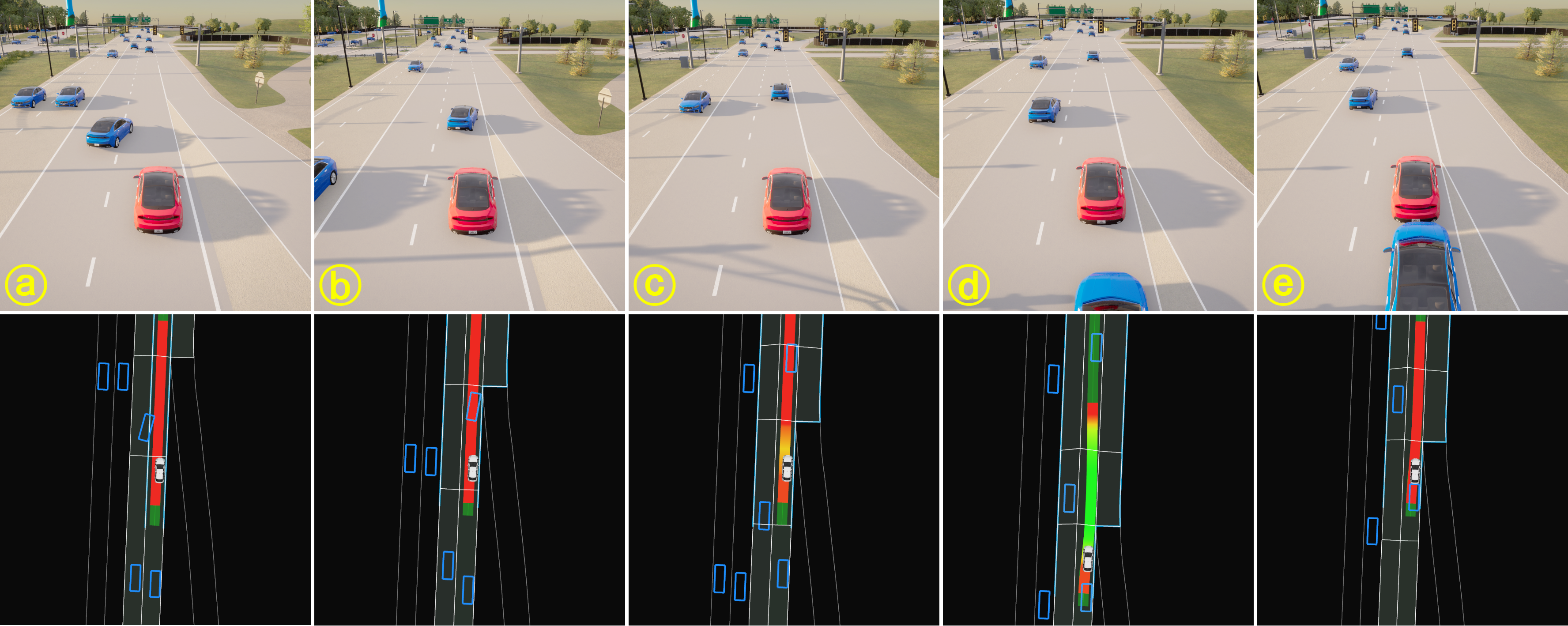}
    \caption{A collision scenario in TeraSim with an Autoware-controlled AV. An aggressive cut-in triggers abrupt braking of AV, followed by cautious acceleration and a rear-end collision. The top row shows CARLA’s car-following view, while the bottom row presents Autoware’s bird’s-eye view. Corresponding columns capture the same timestamp. The sequence unfolds from (a) to (e).}
    \label{fig:autoware}
\end{figure*}

In this section, we present four types of adversities—both static and dynamic—generated by TeraSim. These adversities include construction zones, vehicles, cyclists, and pedestrians, as illustrated in Fig. \ref{fig:adversity}. The red Lincoln MKZ represents the AV, while the surrounding agents are background traffic entities generated by TeraSim.

\textbf{Construction Zone}: The AV drives on a two-lane highway, but the left lane is blocked due to a construction zone with traffic cones, forcing the AV to slow down and merge into the right lane.

\textbf{Vehicle Adversity}: The AV attempts an unprotected left turn at a green light. However, a background vehicle rapidly approaches from the opposite direction on a straight path, requiring the AV to stop and yield.

\textbf{Cyclist Adversity}: The AV shares the road with a cyclist, creating a challenge in deciding whether to follow or overtake while ensuring both efficiency and safety.

\textbf{Pedestrian Adversity}: The AV approaches a small intersection with a green light. However, a pedestrian unexpectedly runs across the road against a red light, forcing the AV to brake and stop.

With these and more adversities, we can generate safety-critical events that are integral to the rigorous testing of AVs. These events help reveal potential vulnerabilities and assess system responses under safety-critical conditions. These valuable events can help build resilience and reliability into AV technology, laying the foundation for safer real-world deployment.

\subsection{Unknown Unsafe Event Demonstration}\label{subsection:case-study}

We evaluated the AV controlled by Autoware Universe, using the proposed simulation system. This evaluation allowed us to successfully identify potential weaknesses, which are illustrated in the following collision example as shown in Fig. \ref{fig:autoware}. The incident resulted from a sequence of adversarial events: an aggressive cut-in maneuver followed by a rear-end collision.

The scenario started when a background vehicle abruptly cut in front of the AV (Fig.~\ref{fig:autoware}a). In response, the AV applied hard braking and came to a complete stop (Fig.~\ref{fig:autoware}b). Once the cut-in vehicle drove away, the AV began recovering but accelerated very cautiously (Fig.~\ref{fig:autoware}c-d). This slow response set the stage for the second adversarial event: the vehicle behind the AV failed to react in time and rear-ended it (Fig. \ref{fig:autoware}e).

Although the AV was not the main cause of the accident, its conservative response significantly contributed to the situation. A more proactive strategy—such as resuming acceleration more quickly or avoiding a complete stop—could have mitigated the risk of a rear-end collision. This case demonstrates TeraSim’s ability to systematically uncover unknown unsafe events that conventional scenario-based testing fails to capture, revealing hidden vulnerabilities in AV decision-making. By exposing AVs to diverse, high-risk interactions that may rarely occur in real-world testing but pose critical safety threats, TeraSim provides a rigorous framework for identifying failure modes and refining AV behaviors to enhance robustness in complex traffic scenarios.

\subsection{Quantitative Safety Assessment}

Unlike traditional scenario-based simulators, which rely on predefined cases and struggle to estimate real-world crash rates, TeraSim constructs a fully dynamic, naturalistic driving environment, enabling quantitative and comprehensive AV safety assessment. For the quantitative analysis, we send the ground truth perception including traffic light status and surrounding objects to test Autoware Universe's planning module using its default simulator. We conducted approximately 6,000 testing episodes over 936 wall-time hours using multiple AWS EC2 C5.9xlarge instances, each equipped with 36 cores and 72 GB RAM. This large-scale evaluation allowed us to quantify the crash rate of Autoware Universe at $4.16 \times 10^{-3}$ crashes per mile, about 2,000 times higher than that of an average human driver ($1.86 \times 10^{-6}$ crash per mile). These results validate TeraSim's ability not only to expose unknown unsafe events, but also to provide crash rate estimation, a critical capability that traditional scenario-based approaches lack.

\vspace{0.1cm}

\section{Conclusion}
In this work, we present TeraSim, an open-source generative traffic simulation framework that integrates a naturalistic and adversarial environment with an adversity-driven paradigm, allowing the discovery of unknown unsafe events and the statistical evaluation of the safety performance of AVs. In addition, we introduce a plugin-based architecture that ensures seamless integration with third-party simulators and AV stacks, fostering broad applicability across different research and development pipelines. 

In the future, we plan to expand TeraSim’s capabilities in two main directions. First, we will integrate large language models into TeraSim, enabling natural language and multi-modal user input for customizing simulations. Second, we plan to increase scalability by developing a macro-micro-integrated simulation framework, which will extend AV testing to city-scale environments while maintaining computational efficiency.

Ultimately, we envision TeraSim as a long-term open-source project that continues to evolve with new functionalities and innovations. By fostering collaboration within the research community, we aim to establish TeraSim as a foundational tool for advancing autonomous vehicle technology development and safety assessment.

\bibliographystyle{ieeetr}
\bibliography{bibtex/bib/terasim}

\end{document}